\documentclass{svjour3}
\usepackage{rotating}
\usepackage{epsfig}
\usepackage{cite}
\usepackage{amssymb}
\usepackage{url}
\usepackage{theorem}
\usepackage[OT1]{fontenc}
\usepackage{float}
\usepackage{here}
\usepackage{lscape}
\usepackage{stfloats}
\usepackage{subfigure}
\usepackage{array}
\usepackage{makeidx}
\usepackage{amstext}
\usepackage{amsfonts}
\usepackage{epsf}
\usepackage{psfrag}
\usepackage[latin1]{inputenc}
\usepackage{cite}
\newcommand{\beqnum}{\begin{equation} \begin{array}{lcl}}
\newcommand{\eeqnum}{\end{array} \end{equation}}
\newcommand{\beqnom}{\begin{eqnarray}}
\newcommand{\eeqnom}{\end{eqnarray}}
\newcommand{\beqnc}{\begin{center}\begin{eqnarray}}
\newcommand{\eeqnc}{\end{eqnarray}\end{center}}
\newcommand{\beqnlm}{\begin{equation}\vspace{-.5cm}\begin{array}{lll}}
\newcommand{\eeqnlm}{\end{array}\end{equation}}\vspace{-.5cm}
\newcommand{\beq}{\begin{eqnarray*}}
\newcommand{\eeq}{\end{eqnarray*}}
\newcommand{\bef}{\begin{figure}}
\newcommand{\enf}{\end{figure}}
\begin{document}

\title{{\itshape Ball on a Beam: Stabilization under Saturated Input Control with Large Basin of Attraction}}
\author{Yannick Aoustin$^\ast$ \thanks{$^\ast$Corresponding author.  \vspace{6pt}\\Yannick
Aoustin\\Institut de
Recherche en Communications et Cybernétique de Nantes U.M.R. 6597\\
1 rue de la No\"e (IRCCyN), BP 92101, 44321 Nantes Cedex 3,
FRANCE\\Email: Yannick.Aoustin@IRCCyN.ec-nantes.fr\\\\Alexander
Formal'skii\\ Institute of Mechanics, Moscow Lomonosov State
University, 1, Michurinskii Prospect, Moscow, 119192, Russia\\Email:
formal@imec.msu.ru}and Alexander Formal'skii \\\vspace{6pt} }
\maketitle

\begin{abstract}
This article is devoted to the stabilization of two  under actuated planar
systems, the well known straight beam-and-ball system and an
original circular beam-and-ball system. The feedback control for
each system is designed, using the Jordan form of its model,
linearized near the unstable equilibrium. The limits on the voltage,
fed to the motor, are taken into account explicitly. The straight
beam-and-ball system has one unstable mode in the motion near the
equilibrium point. The proposed control law ensures that the basin
of attraction coincides with the controllability domain. The
circular beam-and-ball system has two unstable modes near the
equilibrium point. Therefore this device, never considered in the
past, is much more difficult to control than the straight
beam-and-ball system. The main contribution is to propose a simple
new control law, which ensures, by adjusting its gain parameters,
that the basin of attraction arbitrarily can approach the
controllability domain for the linear case. For both nonlinear
systems, simulation results are presented to illustrate the
efficiency of the designed nonlinear control laws and to determine the basin
of attraction.
\end{abstract}


\section{Introduction}
Among the mechanical systems, the under actuated systems, which have
fewer controls than configuration variables, represent a great
challenge for the control.  They are characterized by  the
underactuation degree, which is the difference between the numbers
of configuration variables and controls. An active field of research
exists, due to the applications of under actuated systems such as
aircrafts, satellites with thrusters, spacecrafts, flexible robots,
legged robots, which adopt a dynamical stable walking or running
gait, inverted pendulums. For example, for a planar vertical
take-off and landing aircraft (PVTOL),  an
approximate input-output linearization procedure  is developed in \cite{Hauser92} to get a bounded
tracking and an asymptotic stability. In paper \cite{alimir03}, the stabilization of a satellite is studied when one of its three
thrusters is not efficient. Flexible robots have an infinite number
of flexible modes, which can be damped, using a controller, based on
a discrete model, (see \cite{Book83}, \cite{DeLuca91}, etc...).
Inverted pendulum devices are used like a testbed research or for
education application to investigate new control laws for the
stabilization or the swing up (see \cite{Astrom00}, \cite{grishin02}
or \cite{spong95}). Mechanical models of two planar systems, which
have an unactuated cyclic variable, and all their shape variables
are independently actuated, are considered in
\cite{GrizzleMoChe05}. To deal with the stability of a walking gait
for a biped, which is under actuated in single support, because it
has point feet, in the papers \cite{Aoustin991}, \cite{Aoustin031}
and  \cite{Plestan03}, reference trajectories are defined for the
actuated variables as functions of an undriven strictly monotone
state variable. A complete characterization of all mechanical
systems with underaction degree one is given in
\cite{Acosta04}. In \cite{Fantoni02}, a control law is proposed to
stabilize the surge, sway and angular velocities of the hovercraft
system. We can also note the very interesting thesis document
\cite{Saber010}, which is devoted to nonlinear control, reduction,
and classification of under actuated mechanical systems and in
particular of high order under actuated
 systems. Then numerous mechanical devices associated with
 underactuation have been studied in literature. Furthermore this
 topic is far to be closed, because control design methods do not exist for many under actuated systems that are important for applications.

 This paper deals with the stabilization of two planar under actuated systems.
 The first system is the well-known straight beam-and-ball system. The ball is perfectly rolling without slide on the beam.
 Due to the complexity of this system, the stabilization and the tracking problem using a state or an output feedback
  have been considered by many researchers
 (see \cite{Hauser92_bis}, \cite{TeelPraly95}, \cite{Tellistambul92}, \cite{SepJanKok97} or
\cite{Saber010}). In paper \cite{Hauser92_bis}, tracking for
this system was considered using approximate input-output
linearization.
 Semiglobal stabilization of the straight beam-and-ball system using state feedback was addressed by
\cite{Tellistambul92}. In \cite{TeelPraly95}, this system is
stabilized using output feedback. The problem of global
stabilization of the straight beam-and-ball
 system with friction was considered in  paper  \cite{SepJanKok97}. The viscous friction is taken into account in our paper too.
Semiglobal stabilization of this system, using fixed-point state
feedback was
 addressed by  \cite{Saber010}.

The second system is an original circular beam-and-ball system.
 For each system, a control law, based on the linearized model and
 its Jordan form is designed. The saturation of the
 actuator is taken into account explicitly, so  the control law is non-linear.
 This kind of control has been previously tested
to stabilize a biped with point feet  \cite{Aoustin043},  a one-link pendulum with
flywheel \cite{grishin02},
 and  to stabilize a two-link pendulum with flywheel \cite{Aoustin061}. The main
 difference between the straight beam-and-ball system and the circular beam-and-ball system is that the linear model
  of the second system
 has two
 eigenvalues in the right-half complex plane.
 Therefore, it is more difficult to stabilize the circular beam-and-ball system
 than the  straight beam-and-ball system with only one eigenvalue in the right-half complex plane.
 For the linearized model of the straight beam-and-ball system,
  the controllability domain, noted $Q$, and the basin of attraction, noted $B$, can coincide under a linear control
  law with restriction (see \cite{formal74}, \cite{grishin02}).
  This property
is not satisfied for the linearized model of the circular beam-and-ball system with two
 eigenvalues in the right-half complex plane.  But the basin of attraction $B$  can be made arbitrary close
 to the controllability domain $Q$ (see \cite{HuLinQiu01}).

 We hope that our study is of theoretical interest and also has some
 pedagogical value.

The paper is organized as follows: Section~\ref{BallonStraightBeam}
is devoted to the straight beam-and-ball system. In
Subsection~\ref{BallonStraightBeammodel}, the  equations of motion are
written. The linear model is presented in
Subsection~\ref{BallonStraightBeamlinearizedmodel1}. Subsections
\ref{Controllability1}, \ref{Spectrum1}, \ref{positionpb1}
 are organized to introduce a
control law with saturation, to get a basin of attraction $B$,
which coincides with the controllability domain $Q$. The circular
beam-and-ball system is studied in Section \ref{Round beam-and-ball
system}. In Subsection~\ref{BallonRoundBeammodel}, the
equations of  motion are written. The linear model is presented in
Subsection~\ref{BalloncircleBeamlinearizedmodel}.
Subsections~\ref{Controllability2}, \ref{Spectrum2},
\ref{positionpb2} are organized to introduce a control law with
saturation, to get a large basin of attraction $B$ inside the
controllability domain $Q$. Simulation results for the complete
nonlinear systems are shown to illustrate the efficiency of the
proposed control laws.
 Finally,
Section~\ref{sec:conclusion} contains our conclusion and
perspectives.
\section{Straight beam-and-ball system}\label{BallonStraightBeam}
The straight beam-and-ball system consists of a straight beam and a ball on it, see Figure~\ref{fig:diagram1}. The ball is rolling on the beam without slide. The point $C_{1}$ is center of mass of the beam with its holder $OA$. The point
$C_{2}$ and and value $r$ are center and radius of the ball. The point $C_{2}$ is also the center of mass of the ball.
\subsection{Equations of motion}\label{BallonStraightBeammodel}
Let $m_{1}$ and $m_{2}$  denote the mass of the beam with its holder $OA$ and the mass of the ball, respectively. Let us
 introduce $\rho_{1}$ and $\rho_{2}$ the radii of inertia such
that $I_{1}=m_{1}\rho_{1}^{2}$ and $I_{2}=m_{2}\rho_{2}^{2}$ are respectively the inertia moment of the beam with its holder $OA$
around the suspension point $O$ and the inertia moment of the ball around its center $C_{2}$; let $OC_{1}=a$ and $OA=l$.

Two generalized coordinates, the angular variables $\theta$ and
$\varphi$ characterize the behavior of this system. Position of the
ball on the beam is defined also by the distance $s=r \varphi$. Let
$\Gamma$ be the torque, which is directly proportional to the
electrical current in the armature winding. By neglecting the
armature inductance (in other words, the electromagnetic time
constant in the rotor circuit), this torque can be written in the
form (see \cite{Gorinevsky97}):
\begin{equation}\label{eq:torque}
    \Gamma=c_{u}u-c_{v}\dot{\theta}
\end{equation}
where $u$ is the voltage, supplied to the motor. The positive
constants $c_{u}$ and $c_{v}$ for a given motor can be calculated by
using the values for the starting torque, the nominal voltage, the
nominal torque and the nominal angular velocity \cite{Gorinevsky97}.
Product $c_{v}\dot{\theta}$ is the torque of the back
electromotive force. The torque of the viscous friction force in the joint $O$ (if it is taken into account) is also proportional to angular velocity $\dot{\theta}$. We will  consider the following constraint,
imposed on  the voltage $u$:
    \begin{equation}\label{eq:voltagemax}
    |u|\,\leq\,u_{0}, \,\thinspace
    \thinspace u_{0}\,=\,const
\end{equation}

 The expressions for the kinetic energy $K$ and the potential energy
 $\Pi$ are the following ($g$ is
 the gravity acceleration):
 \begin{equation}\label{eq:kineticenergy1}
 \begin{array}{c}
   2K=m_{1}\rho_{1}^{2}\dot{\theta}^{2}+m_{2}[r^{2}\varphi^{2}+(r+l)^{2}]\dot{\theta}^{2}+2m_{2}r(l+r)\dot{\varphi}\dot{\theta}+m_{2}(r^{2}+\rho_{2}^{2})\dot{\varphi}^{2}
   \\\\
   \Pi=m_{1}gacos\theta+m_{2}g[-r\varphi sin\theta + (l+r)cos\theta]
 \end{array}
 \end{equation}
The equations of the mechanism motion can be derived, using
Lagrange's method:
 \begin{equation}\label{eq:motionequation1}
 \begin{array}{c}
   \left[m_{1}\rho_{1}^{2}+m_{2}(r+l)^{2}+m_{2}r^{2}\varphi^{2}\right]\ddot{\theta}+m_{2}r(r+l)\ddot{\varphi}+2m_{2}r^{2}\varphi\dot{\varphi}\dot{\theta}-\\
-g[m_{1}a+m_{2}(r+l)]sin\theta-m_{2}gr\varphi
cos\theta=c_{u}u-c_{v}\dot{\theta}
 \end{array}
 \end{equation}

\begin{equation}\label{eq:motionequation2}
    r(r+l)\ddot{\theta}+(r^{2}+\rho_{2}^{2})\ddot{\varphi}-r^{2}\varphi\dot{\theta}^{2}-grsin\theta=0
\end{equation}
 If $u=0$,  system
(\ref{eq:motionequation1}), (\ref{eq:motionequation2}) has one
unstable equilibrium state:
\begin{equation}\label{eq:equilibrium1}
    \theta=0,\quad \varphi=0\quad (s=0),\quad \dot{\theta}=0,\quad \dot{\varphi}=0\quad (\dot{s}=0)
\end{equation}

\subsection{Linearized Model}\label{BallonStraightBeamlinearizedmodel1}
Corresponding to the nonlinear equations
(\ref{eq:motionequation1}),~(\ref{eq:motionequation2}), the linear
model of the motion near the unstable equilibrium state
(\ref{eq:equilibrium1}) is:
\begin{equation}\label{eq:linear1}
      \left[m_{1}\rho_{1}^{2}+m_{2}(r+l)^{2}\right]\ddot{\theta}+m_{2}r(r+l)\ddot{\varphi}
-g[m_{1}a
+m_{2}(r+l)]\theta-m_{2}gr\varphi=c_{u}u-c_{v}\dot{\theta} \\
\end{equation}
\begin{equation}\label{eq:linear2}
     r(r+l)\ddot{\theta}+(r^{2}+\rho_{2}^{2})\ddot{\varphi}-gr\theta=0
\end{equation}

\subsection{Kalman controllability}\label{Controllability1}
The determinant of the controllability matrix (see \cite{kalman69})
for the linear model (\ref{eq:linear1}), (\ref{eq:linear2}) is not null,
if and only if:

 \begin{equation}\label{eq:condicom1}
    r^{2}g^{2}\left[(2r^{2}+\rho_{2}^{2})\rho_{2}^{2}+r^{4}\right]\neq0
 \end{equation}
Thus,  inequality (\ref{eq:condicom1}) is valid, if $r\neq0$. If $r=0$, then the ball becomes a material point and we do not consider this case.
Thus, the linear model of the straight beam-and-ball system is
always controllable.

\subsection{Spectrum of Linear System}\label{Spectrum1}
The state form of  system (\ref{eq:linear1}),~(\ref{eq:linear2}), using the state vector $x=(\theta,~\varphi,~\dot{\theta},~\dot{\varphi})^{T}$, is:
\begin{equation}\label{eq:stateform1}
    \dot{x}=Ax+bu=\left[
              \begin{array}{cc}
                0_{2\times2}~ & ~ I_{2\times2} \\\\
                D^{-1}E ~ & ~ D^{-1}\left(
                                  \begin{array}{cc}
                                    -c_{v}& 0 \\
                                    0 & 0 \\
                                  \end{array}
                                \right)
              \end{array}
            \right]x+ \left[\begin{array}{c}
                                             0 \\
                                             0 \\
                                             D^{-1}\left(
                                                     \begin{array}{c}
                                                       c_{u} \\
                                                       0 \\
                                                     \end{array}
                                                   \right)
                                              \\
                                           \end{array}\right]u
\end{equation}
The notations $0_{2\times2}$ and $I_{2\times2}$ define a zero matrix and an identity matrix, respectively. The expressions of matrices $D$ and $E$ are
\begin{equation}\label{eq:matricesDB}
\begin{array}{c}
      D=\left(
        \begin{array}{cc}
          m_{1}\rho_{1}^{2}+m_{2}(r+l)^{2}~ & ~ m_{2}r(r+l) \\
          r(r+l)~ & ~ r^{2}+\rho_{2}^{2} \\
        \end{array}
      \right) \\\\
      E=g\left(
        \begin{array}{cc}
          m_{1}a+m_{2}(r+l)~ & ~ m_{2}r \\
          r~ & ~ 0 \\
        \end{array}
      \right)
\end{array}
\end{equation}

Introducing a nondegenerate linear transformation $x=Sy$ with a
constant matrix $S$, it is possible to get the well-known Jordan
form of the matrix equation (\ref{eq:stateform1})
 \begin{equation}\label{eq:equajordan1}
    \dot{y}=\Lambda y+du
\end{equation}
where
  \begin{equation}\label{eq:eigenvalmat1}
  \begin{array}{c}
        \Lambda =S^{-1}AS=\left(%
\begin{array}{ccccc}
  \lambda_{1} &   &      & 0 \\
    & \lambda_{2} &   &      \\
    &   & \lambda_{3} &      \\
  0 &   &      & \lambda_{4} \\
\end{array}%
\right), \quad
    d=S^{-1}b=[d_{i}]^{T}\,\quad (i=1,...,4). \\
  \end{array}
\end{equation}
Here, $\lambda_{1},...,\lambda_{4}$ are the eigenvalues of  the matrix $A$. They are the roots of the characteristic equation of  system
(\ref{eq:linear1}),~(\ref{eq:linear2}):
\begin{equation}\label{eq:polyalex}
    a_{0}\lambda^{4} + a_{1}\lambda^{3} + a_{2}\lambda^{2} + a_{3}\lambda + a_{4}=0
\end{equation}
with
 \\\\$ a_{0}=detD>0,\quad a_{1}=c_{v}(r^{2}+\rho_{2}^{2})>0,\quad
    a_{2}=m_{2}(r+l)(r^{2}-\rho_{2}^{2})-m_{1}a(r^{2}+\rho_{2}^{2}),\\\\ a_{3}=0$,\quad $ a_{4}=detE=-m_{2}g^{2}r^{2}<0.\\\\$
If all physical parameters of the studied system are known, matrix
$S$ of the transformation $x=Sy$ can be calculated.

 According to the theorem of Routh-Hurwitz (see \cite{GraninoTheresa68}), equation (\ref{eq:polyalex}) has one root in the right-half complex plane and three roots in the left-half complex plane (see also
\cite{Saber010}). This assertion does not depend on the sign of the
coefficient $a_{2}$. Of course, the unique root in the right-half
complex plane is located on the real axis.

In Section~\ref{Round beam-and-ball system}, we consider the ball on the circular beam and use linearized model in the same matrix form (\ref{eq:stateform1}) as for the ball on the straight beam, but with different submatrices $D$ and $E$.

\subsection{Problem Statement}\label{positionpb1}
Let $x=0$   (here $0$ is a ($4 \times 1$) zero-column) be the desired equilibrium state of  system (\ref{eq:stateform1}).
 Let us design the feedback control $u(x)$ to stabilize this equilibrium state $x=0$, under constraint (\ref{eq:voltagemax}).
 In other words, we want to design an admissible (satisfying the inequality (\ref{eq:voltagemax}))
  feedback control $|u(x)|\leq u_{0}$ to ensure the asymptotic stability of the desired state $x=0$.
    Let $W$ be the set of piecewise continuous functions of time $u(t)$, satisfying inequality (\ref{eq:voltagemax}).
    Let $Q$ be the set of the initial
     states $x(0)$ of  system (\ref{eq:stateform1}), from which origin $x=0$  can be reached, using admissible control functions of time $u(t)$.
     In other words,  system (\ref{eq:stateform1}) can reach  the origin $x=0$  with the control $u(t)\,\in\,W$, only starting from the initial states $x(0)\,\in\,Q$.
     Set $Q$ is called controllability domain. If the matrix $A$ has eigenvalues with positive real parts and the control
     variable $u$ is restricted, then
     the controllability domain $Q$ for  system (\ref{eq:stateform1}) is an open subset of the phase space $X$ (see
     \cite{formal74}, \cite{grishin02}).

For any admissible feedback
     control $u=u(x)$ with  saturation $|u(x)|\,\leq\,u_{0}$ the corresponding basin of attraction belongs to
     the controllability domain: $B \subset Q$. Here, as usual, $B$  is the set of initial states
     $x(0)$, from which system (\ref{eq:stateform1}), with  feedback
$u=u(x)$ asymptotically tends to the origin point $x=0$ as
     $t \rightarrow \infty$.

     In the following section, a control law will be presented for
     the straight beam-and-ball system to get a basin of attraction $B$, which coincides with
     the controllability domain $Q$:
     $B=Q$.
     \subsection{Feedback Control for the straight beam-and-ball
     system}\label{control1}
     A control law is proposed here to stabilize  the straight beam-and-ball
     system with basin of attraction as large as possible.
     \subsubsection{Control design}
Let $\lambda_{1}$ be the real positive eigenvalue, $Re\lambda_{i}<0$
($i=2,~3,~4$) and let us consider the first scalar differential
equation of system (\ref{eq:equajordan1}) corresponding to
eigenvalue $\lambda_{1}$,
     \begin{equation}\label{eq:scalary1}
    \dot{y}_{1}=\lambda_{1}y_{1}+d_{1}u
\end{equation}
    System (\ref{eq:stateform1}),  is a Kalman controllable system, therefore  scalar $d_{1}\neq0$. The controllability domain $Q$ of the equation (\ref{eq:scalary1}) and consequently of  system (\ref{eq:equajordan1}) is described by the following inequality
   (see \cite{formal74}, \cite{grishin02})
\begin{equation}\label{eq:contrailimit}
    \left|y_{1}\right|<\left|d_{1}\right|\frac{u_{0}}{\lambda_{1}}
\end{equation}

 The instability of the  coordinate $y_{1}$ can be ``suppressed'' by a linear feedback control,
 \begin{equation}\label{eq:feedlin}
    u=\gamma y_{1}
\end{equation}
with the following condition,
        \begin{equation}\label{eq:condition2}
    \lambda_{1}+d_{1}\gamma\,<\,0
\end{equation}

For  system (\ref{eq:stateform1}) under the feedback control
(\ref{eq:feedlin}) with  inequality (\ref{eq:condition2}), only
the pole $\lambda_{1}$ is replaced by a negative pole
$\lambda_{1}+d_{1}\gamma$. The poles $\lambda_{2}$, $\lambda_{3}$,
$\lambda_{4}$ do not change.

    If  constraint (\ref{eq:voltagemax}) is taken into account, the linear feedback control (\ref{eq:feedlin})  becomes with saturation,
\begin{equation}\label{eq:solrealbi2}
u=u(y_{1})=
\left\{%
\begin{array}{rlll}
  u_{0}, & \quad if & \gamma y_{1}\geq u_{0} &  \\
  \\
  \gamma y_{1}, & \quad if & |\gamma y_{1}|\leq
     u_{0} &  \\
     \\
  -u_{0}, & \quad if & \gamma y_{1} \leq -u_{0} & \\
\end{array}%
\right.
\end{equation}

The unit of coefficient $\gamma$ is volt.

    It is possible to see that if $|y_{1}|<|d_{1}|u_{0}/\lambda_{1}$,
    then under  condition (\ref{eq:condition2}) the right part of  equation (\ref{eq:scalary1})
    with the nonlinear control (\ref{eq:solrealbi2}) is negative
     when $y_{1}>0$  and positive when $y_{1}<0$. Consequently, if $|y_{1}(0)|<|d_{1}|u_{0}/\lambda_{1}$,
     then  the solution $y_{1}(t)$ of  system (\ref{eq:scalary1}), (\ref{eq:solrealbi2}) tends to 0  as $t \rightarrow
     \infty$. But if $y_{1}(t) \rightarrow 0$, therefore, according to  expression
(\ref{eq:solrealbi2}), $u(t)\rightarrow 0$ as $t \rightarrow \infty$. Therefore,   the solutions $y_{i}(t)$
      ($i=2, 3, 4$) of the second, third and fourth equations of  system (\ref{eq:equajordan1}) with any initial
      conditions $y_{i}(0)$
         ($i=2, 3, 4$) converge to zero as $t \rightarrow \infty$, because $Re\lambda_{i}<0$  for $i=2, 3, 4$.
         Thus, under the nonlinear control (\ref{eq:solrealbi2}) and
         with  inequality (\ref{eq:condition2}), the basin of attraction $B$  coincides with the controllability domain $Q$
           (see \cite{formal74}, \cite{grishin02}): $B=Q$. So, the basin of attraction  $B$  for  system (\ref{eq:stateform1}), (\ref{eq:solrealbi2})
          is as large as possible and it is
          described by  inequality (\ref{eq:contrailimit}).

    Note that  the variable $y_{1}$ depends on the original variables from  the vector $x$, according to  the transformation $x=Sy$
     or $y=S^{-1}x$.
    Due to this, formula (\ref{eq:solrealbi2}) defines the control feedback, which depends on  the vector $x$ of the original variables. If the matrix $S$ is calculated, then all coefficients of the designed control can be defined. Only the constant $\gamma$ is an arbitrary multiplier, but it has to satisfy  inequality (\ref{eq:condition2})

      Thus, linearizing nonlinear system (\ref{eq:motionequation1}), (\ref{eq:motionequation2}), (\ref{eq:solrealbi2})  near the equilibrium state we obtain a system, which is asymptotically stable. Using Lyapounov's theorem (see \cite{KHALIL02}), we conclude that  equilibrium (\ref{eq:equilibrium1}) of the nonlinear system (\ref{eq:motionequation1}), (\ref{eq:motionequation2}) is asymptotically stable under
     control (\ref{eq:solrealbi2}) with some basin of attraction. In the next Subsection, numerically we  find the upper bounds of the initial values of some variables, which can be handled for the linear and nonlinear models under the designed control.

\subsubsection{Numerical results}\label{sectionnum}
Let
\begin{equation}\label{eq:data1}
  \begin{array}{c}
    m_{1}=1.0\thinspace kg,\quad m_{2}=0.2\thinspace kg, \quad g=9.81\thinspace m/s^{2},\\
    \\
    r=0.05\thinspace m,\quad
     l=0.2\thinspace m,\quad
     a=0.15\thinspace m,\quad
    \rho_{1}=0.2179\thinspace
    m,\quad \rho_{2}=0.1414\thinspace m,\thinspace\\ \\
c_{u}=0.007\thinspace N\raisebox{0.08cm}{.}m/V, \quad
c_{v}=0.0001\thinspace N\raisebox{0.08cm}{.}m/s, \quad u_{0}=19\thinspace V.\\
  \end{array}
\end{equation}
In open-loop the poles of the linear system (\ref{eq:stateform1}) (the roots of  equation (\ref{eq:polyalex})) with  parameters (\ref{eq:data1}) are:
\begin{equation}\label{eq:eigenvalue1}
      \lambda_{1}=5.7202,\quad \lambda_{2}=-5.7218,\quad \lambda_{3},\lambda_{4}=-2.8\raisebox{0.08cm}{.}10^{-7}\pm1.0558i,\quad  \\
\end{equation}

Now we can use  inequality (\ref{eq:contrailimit}) to evaluate
the basin of attraction $B$  for  system (\ref{eq:stateform1}),
(\ref{eq:solrealbi2}). If
$\theta(0)=\dot{\theta}(0)=\dot{\varphi}(0)=0$, the upper bound of
the initial angles $\varphi$, which can be handled for the linear
model (\ref{eq:stateform1}) is  $\varphi(0)\cong77.679°$. The
corresponding initial distance $s(0)$ is equal to $0.0678~m$. This
value for  the distance $s$ is close to the value
\begin{equation}\label{eq:egali_s}
    s(0)=\frac{c_{u}u_{0}}{m_{2}g}
\end{equation}
With $\theta=0$   product $sm_{2}g$ is the torque about joint O
of the gravity force of the ball (see the nonlinear equations
(\ref{eq:motionequation1}), (\ref{eq:motionequation2}) and the
linear equations (\ref{eq:linear1}), (\ref{eq:linear2})), the
product $c_{u}u_{0}$ is the torque (maximal as possible) developed
by the motor in static. Thus, the point
\begin{equation}\label{eq:condition}
    \theta=\dot{\theta}=\dot{s}=0,\quad s=\frac{c_{u}u_{0}}{m_{2}g}
\end{equation}
is the equilibrium state (unstable) for our system (nonlinear
(\ref{eq:motionequation1}), (\ref{eq:motionequation2}) and linear
(\ref{eq:linear1}), (\ref{eq:linear2})). It is easily to see that
the equilibrium point (\ref{eq:condition}) is located on the
boundary of the controllability region (\ref{eq:contrailimit}).
Simulation shows that, if
\begin{equation}\label{eq:straitbeam}
\theta(0)=\dot{\theta}(0)=\dot{s}(0)=0, \quad s(0)\geq\frac{c_{u}u_{0}}{m_{2}g},
\end{equation}
then it is not possible to
bring the nonlinear system (\ref{eq:motionequation1}),
(\ref{eq:motionequation2}) under control (\ref{eq:solrealbi2}) to the equilibrium
(\ref{eq:equilibrium1}); but it is possible to do that, if
$s(0)<\frac{c_{u}u_{0}}{m_{2}g}$. Furthermore, we think there is no an admissible control $|u(x)|\leq u_{0}$ to bring  system (\ref{eq:motionequation1}), (\ref{eq:motionequation2}) to the equilibrium point (\ref{eq:equilibrium1}) from the initial states (\ref{eq:straitbeam}). This opinion is based on the numerical studies and physical feeling. We do not prove here corresponding assertion strictly.

The eigenvalues $\lambda_{3}$, $\lambda_{4}$ are very close to the
imaginary axis (see (\ref{eq:eigenvalue1})) and therefore under the
control (\ref{eq:solrealbi2}), the transient process is very long.
Let us take into account a viscous friction in  the joint $O$ defined by the  torque
$f\dot{\theta}$. The consideration of the  torque
$f\dot{\theta}$ of the friction force is equivalent to the
consideration in equation (\ref{eq:torque}) of the  term
$(c_{v}+f)\dot{\theta}$ instead of  the term $c_{v}\dot{\theta}$.
With $f=0.4~N\raisebox{0.08cm}{.}m\raisebox{0.08cm}{.}s$ for example
the poles of the corresponding linear system (\ref{eq:stateform1})
in open-loop are:
\begin{equation}\label{eq:eigenvalue2}
      \lambda_{1}=3.4001,\quad \lambda_{2}=-10.0181,\quad \lambda_{3},\lambda_{4}=-0.1041\pm1.0297i \\
\end{equation}
The technique of the feedback control design with a viscous friction
(with new poles (\ref{eq:eigenvalue2})) remains the same exactly.
And the structure of this control remains the same -
(\ref{eq:solrealbi2}). Under the control law (\ref{eq:solrealbi2})
with new coefficients, the transient process converges to the
equilibrium state (\ref{eq:equilibrium1}) faster than without
friction. Using  inequality (\ref{eq:contrailimit}), or the
equality (\ref{eq:egali_s}) we get of course the same value
$s(0)$ as above without friction. So, we can use  formula (\ref{eq:egali_s}) for the linear and
nonlinear systems to calculate  the upper bound of the initial
distances $s$, which are possible
 to stabilize  the equilibrium state (\ref{eq:equilibrium1}).

 Figures~\ref{fig:phipost1} and
\ref{fig:voltage1post1} show a numerical test with an initial tilt
$\varphi(0)=77.65°$ for the nonlinear system
(\ref{eq:motionequation1}), (\ref{eq:motionequation2}) with the
coefficient $f=0.4~N\raisebox{0.08cm}{.}m\raisebox{0.08cm}{.}s$
under the control law (\ref{eq:solrealbi2}) with $\gamma=-122$.
 The voltage, supplied to the motor, is shown in Figure
 \ref{fig:voltage1post1}. The limit value $u_{0}=-19~V$ is
 reached at initial time.

Let $F$ be the reaction force, applied to the ball  orthogonally to
the beam in their contact point. The following formula for this
force holds:
\begin{equation}\label{eq:force1}
    F=m_{2}\left[g~cos\theta-(l+r)\dot{\theta}^{2}-2r\dot{\varphi}\dot{\theta}-r\varphi\ddot{\theta}\right]
\end{equation}
If the reaction force $F$ becomes negative, then the ball loses contact with the beam and our model (with contact) becomes false to describe the physical process.
In the numerical experiment, presented in Figures~\ref{fig:phipost1} and \ref{fig:voltage1post1},  the force $F$ is always positive. This force is shown in
Figure~\ref{fig:force1}.

If $\varphi(0)=\dot{\varphi}(0)=\dot{\theta}(0)=0$, then, using
inequality (\ref{eq:contrailimit}), the upper bound of the initial
tilts of the beam, which can be handled, for the linear model
(\ref{eq:stateform1}) with the friction is $\theta(0)=3.61°$. The
computations show that the upper bound of the initial tilts for the
nonlinear system (\ref{eq:motionequation1}),
(\ref{eq:motionequation2}) under  control (\ref{eq:solrealbi2})
is $\theta(0)=3.64°$. So, this value is little more important than
for the linear system (\ref{eq:stateform1}) under the same control
(\ref{eq:solrealbi2}).
\section{Circular beam-and-ball system}\label{Round beam-and-ball system}
The circular beam-and-ball system consists of a circular beam with  the center $C$ and   the radius $R$ and a ball on it with  the center $C_{2}$ and  the radius
$r$, see Figure~\ref{fig:diagram2}. The point $C_{1}$ is  the center of mass of  the beam with its holder $OA$.


\subsection{Equations of motion}\label{BallonRoundBeammodel}
Here  the same notations are used, that for the straight beam-and-ball system.

Let $m_{1}$ and $m_{2}$  denote  the mass of the beam with its holder $OA$ and the mass of the ball, respectively. Let $\rho_{1}$ and $\rho_{2}$ be the radii
of inertia respectively of the beam with its holder $OA$ and of the ball; let $OC_{1}=a$ and $OA=l$ be.

The generalized coordinates are the joint variable $\theta$ and the
angle variable $\varphi$. Position of the ball on the beam is also
defined by  distance $s=r \varphi$. The relation between the
angle $\varphi$ and  angle $\psi$ is:
\[
    r\varphi=R\psi
\]

Let us assume that the motor is the same that for the straight beam-and-ball system with  torque (\ref{eq:torque}) and
 constraint (\ref{eq:voltagemax}).
The constants $c_{u}$, $c_{v}$ and $u_{0}$ are the same.

The expressions for the kinetic energy $K$ and the potential energy
 $\Pi$ are the following:
 \begin{equation}\label{eq:kineticenergy}
 \begin{array}{c}
   2K=\left\{m_{1}\rho_{1}^{2}+m_{2}\left[(R+r)^2+(l-R)^2+2(R+r)(l-R)cos\frac{r\varphi}{R}\right]\right\}\dot{\theta}^{2}+\\
   +m_{2}\left[(R+r)^2+\frac{(\rho_{2}R)^{2}}{r^{2}}\right]\frac{(r\dot{\varphi})^{2}}{R^{2}}+\\
   +2m_{2}\left[(R+r)^2+(R+r)(l-R)cos \frac{r\varphi}{R}\right]\dot{\theta}\frac{r\dot{\varphi}}{R}\\\\
   \Pi=[m_{1}a+m_{2}(l-R)]gcos\theta+m_{2}g(R+r)cos(\frac{r}{R}\varphi+\theta)
 \end{array}
 \end{equation}
The equations of the mechanism motion are derived, using Lagrange's method:
 \begin{equation}\label{eq:motionequation3}
 \begin{array}{c}
 \left[m_{1}\rho_{1}^{2}+m_{2}(r^{2}+l^{2}+2rlcos\frac{r\varphi}{R})+2m_{2}R(R+r-l)(1-cos\frac{r\varphi}{R})\right]\thinspace\ddot{\theta}+\\\\
 +m_2r(1+\frac{r}{R})\left[R+r+(l-R)cos\frac{r\varphi}{R}\right]\thinspace\ddot{\varphi}+m_2r(1+\frac{r}{R})(R-l)(2\dot{\theta}+
 \frac{r\dot{\varphi}}{R})\dot{\varphi}sin\frac{r\varphi}{R}-\\\\
-g[m_{1}a\thinspace+m_{2}(l-R)]sin\theta-m_{2}g(R+r)sin(\theta+\frac{r\varphi}{R})=c_{u}u-c_{v}\dot{\theta}
 \end{array}
 \end{equation}
 \\
 \begin{equation}\label{eq:motionequation4}
 \begin{array}{c}
r(1+\frac{r}{R})\left[R+r+(l-R)cos\frac{r\varphi}{R}\right]\thinspace\ddot{\theta}+\left[\rho_{2}^{2}+r^{2}(1+\frac{r}{R})^{2}\right]\thinspace\ddot{\varphi}+\\\\
+(1+\frac{r}{R})(l-R)\dot{\theta}^{2}sin\frac{r\varphi}{R}-gr(1+\frac{r}{R})sin(\theta+\frac{r\varphi}{R})=0
 \end{array}
 \end{equation}
 \\
 If $u=0$,  system (\ref{eq:motionequation3}),~(\ref{eq:motionequation4}) has
one unstable equilibrium state (\ref{eq:equilibrium1}).

\subsection{Linearized Model} \label{BalloncircleBeamlinearizedmodel}
Linearizing the equations (\ref{eq:motionequation3}),~(\ref{eq:motionequation4}) near the unstable equilibrium state (\ref{eq:equilibrium1}),  we get the
following model:
 \begin{equation}\label{eq:linear3}
 \begin{array}{c}
 \left[m_{1}\rho_{1}^{2}+m_{2}(r+l)^{2}\right]\ddot{\theta}+
 +m_2r\left(1+\frac{r}{R}\right)(r+l)\ddot{\varphi}-\\\\
-g[m_{1}a+m_{2}(r+l)]\theta-m_{2}g(r+R)\frac{r\varphi}{R}=c_{u}u-c_{v}\dot{\theta}
 \end{array}
 \end{equation}
 \\
  \begin{equation}\label{eq:linear4}
 \begin{array}{c}
r\left(1+\frac{r}{R}\right)(r+l)\thinspace\ddot{\theta}+\left[\rho_{2}^{2}+r^{2}\left(1+\frac{r}{R}\right)^{2}\right]\thinspace\ddot{\varphi}
-gr\left(1+\frac{r}{R}\right)\left(\theta+\frac{r\varphi}{R}\right)=0
 \end{array}
 \end{equation}
\subsection{Kalman controllability}\label{Controllability2}
The determinant of the controllability matrix  for  the model (\ref{eq:linear3}),~(\ref{eq:linear4}) is not null, if and only if:

 \begin{equation}\label{eq:condicom2}
   Rr^{2}(R-l)+R^{2}\rho_{2}^{2}+r^{3}(R-l)\neq0
 \end{equation}

If $r=0$, then the ball becomes a material point and $\rho_{2}=0$. In this case, instead of inequality (\ref{eq:condicom2}) the equality is correct. However, we do not consider a material point on the beam and therefore assume $r\neq0$.

Let $r\neq0$, but the mass of the ball is concentrated in its center
$(\rho_{2}=0)$ and the suspension point $O$ coincides with the
curvature center $C$ of the circular beam ($R=l$). In this case,
inequality (\ref{eq:condicom2}) is not satisfied and the linear
system is not controllable. Consider the controllability of the
original nonlinear system
(\ref{eq:motionequation3}),~(\ref{eq:motionequation4}) in the case
$\rho_{2}=0$ and $R=l$. Introduce the angle
$\alpha=\theta+\frac{r\varphi}{R}$. The nonlinear system
(\ref{eq:motionequation3}),~(\ref{eq:motionequation4}) becomes:
 \begin{equation}\label{eq:nonlinamatpoin3}
 \begin{array}{c}
m_{1}\rho_{1}^{2}\ddot{\theta} - m_{1}ga sin\theta=c_{u}u-c_{v}\dot{\theta}
 \end{array}
 \end{equation}
\begin{equation}\label{eq:nonlinamatpoin4}
   (R+r)\ddot{\alpha}-gsin\alpha=0
\end{equation}
The equations (\ref{eq:nonlinamatpoin3}) and
(\ref{eq:nonlinamatpoin4}) are separated. The control $u$ has no
action on   the angle $\alpha$ and  system
(\ref{eq:nonlinamatpoin3}), (\ref{eq:nonlinamatpoin4}) is not
controllable.

Inequality (\ref{eq:condicom2}) is satisfied, if
\begin{equation}\label{eq:inequalitycircular}
    l-R\neq\frac{R^{2}\rho_{2}^{2}}{(R+r)r^{2}}
\end{equation}
and we will consider only this case.
\subsection{Spectrum of Linear System}\label{Spectrum2}
The state form of  system (\ref{eq:linear3}),~(\ref{eq:linear4}) can be presented in the same matrix form (\ref{eq:stateform1}) as for the straight beam, but with the following submatrices $D$ and $E$:
\begin{equation}\label{eq:matricesDB}
\begin{array}{c}
      D=\left(
        \begin{array}{cc}
         m_{1}\rho_{1}^{2}+m_{2}(r+l)^{2}~&~m_2r(1+\frac{r}{R})(r+l)
         \\\\
         r(1+\frac{r}{R})(r+l)~&~ \rho_{2}^{2}+r^{2}(1+\frac{r}{R})^2 \\
        \end{array}
      \right) \\\\
      E=g\left(
        \begin{array}{cc}
          m_{1}a
+m_{2}(r+l)~ & ~m_{2}r(1+\frac{r}{R}) \\\\
          r(1+\frac{r}{R})~ & ~ r(1+\frac{r}{R})\frac{r}{R} \\
        \end{array}
      \right)
\end{array}
\end{equation}

Introducing a nondegenerate linear transformation $x=Sy$ with a
constant matrix $S$,  we can get the Jordan form similar to
(\ref{eq:equajordan1}), (\ref{eq:eigenvalmat1}).

The characteristic equation  of  system (\ref{eq:linear3}), (\ref{eq:linear4})  has  form (\ref{eq:polyalex}) with
\\
 \\$ a_{0}=detD>0,\quad a_{1}=c_{v}\left[\rho_{2}^{2}+\frac{r^{2}}{R^{2}}(R+r)^{2}\right]>0,\quad\\\\
 a_{2}=-m_{1}g\left[(\rho_{2}^{2}+r^{2})aR^{2}+(2ar+\rho_{1}^{2})Rr^{2}+(ar+\rho_{1}^{2})r^{3}\right]- \\\\
 ~~~~~~~-m_{2}g\left[(r+l)R^{2}(r^{2}-\rho_{2}^{2})+(r^{2}-l^{2})r^{2}R-(r+l)lr^{3}\right], \\\\
     a_{3}=-c_{v}g\frac{r^{2}}{R^{2}}(R+r)<0,\quad$ $ a_{4}=detE=g^{2}\frac{r^{2}}{R^{2}}(R+r)[m_{1}a+m_{2}(l-R)]. \\$

We assume that
   \begin{equation}\label{eq:inegalitecercle}
    m_{1}a+m_{2}(l-R)>0
   \end{equation}
Inequality (\ref{eq:inegalitecercle}) is satisfied, if  the radius $R$ of the circular beam is sufficiently small (the curvature of the beam is sufficiently large). But we have not to forget  condition (\ref{eq:condicom2}) (or (\ref{eq:inequalitycircular})) of controllability.

Under  condition (\ref{eq:inegalitecercle}),  the coefficient
$a_{4}$ is positive. Using the theorem of Routh-Hurwitz (see
\cite{GraninoTheresa68}), we can conclude that the characteristic
equation (\ref{eq:polyalex}) has two roots in the right-half complex
plane and two roots in the left-half complex plane. This conclusion
does not depend on the sign of  the  coefficient $a_{2}$.

\subsection{Problem Statement}\label{positionpb2}
We will consider the same problem, as before for the straight beam-and-ball system. We want to design an admissible (satisfying  inequality
(\ref{eq:voltagemax})) feedback control to ensure the asymptotic stability of   the state $x=0$ with a large basin of attraction for this equilibrium state.

\subsection{Feedback control for the circular beam-and-ball system}\label{control2}
     A feedback control law $u(x)$, satisfying  inequality (\ref{eq:voltagemax}), is proposed here to stabilize
     the circular beam-and-ball
     system with a large basin of attraction. Under condition (\ref{eq:inegalitecercle}), the linear model of the system has two eigenvalues
$\lambda_{1}$, $\lambda_{2}$ in the right-half complex plane and two eigenvalues $\lambda_{3}$, $\lambda_{4}$ in the left-half complex plane.
\subsubsection{Control design}
 Let $\lambda_{1}$ and $\lambda_{2}$ be the real positive eigenvalues, and let us consider the first two scalar differential equations of  system
(\ref{eq:equajordan1}), (\ref{eq:eigenvalmat1}) for the circular beam, corresponding to
these eigenvalues $\lambda_{1}$ and $\lambda_{2}$:
\begin{equation}\label{eq:primesjor2}
        \dot{y}_{1}=\lambda_{1}y_{1}+d_{1}u,\quad
       \dot{y}_{2}=\lambda_{2}y_{2}+d_{2}u
\end{equation}

Under condition (\ref{eq:inequalitycircular})  system (\ref{eq:stateform1}) for the circular beam is Kalman controllable. Therefore,
 subsystem (\ref{eq:primesjor2}) is controllable too (see
\cite{kalman69}) and $d_{1}\neq0$, $d_{2}\neq0$. The controllability
domain $Q$ of the equations (\ref{eq:primesjor2}), and consequently
of  system (\ref{eq:equajordan1}), is an open bounded set with
the following boundaries (see \cite{Boltyansky66})
\begin{equation}\label{eq:boudaries_rect}
\begin{array}{l}
  y_{1}(\tau)=\pm\frac{d_{1}u_{0}}{\lambda_{1}}\left(2e^{-\lambda_{1}\tau}-1\right),\\\\
  y_{2}(\tau)=\pm\frac{d_{2}u_{0}}{\lambda_{2}}\left(2e^{-\lambda_{2}\tau}-1\right)\quad
   \left(0\leq\tau<\infty\right)
\end{array}
\end{equation}

 If the
system has two complex poles in the right-half complex plane, then
instead of (\ref{eq:boudaries_rect}) we will get other formulas
(see \cite{formal74}).

Set $Q$ belongs to the rectangle, defined by
inequalities:\\\\
    \centerline{$|y_{1}|<|d_{1}|\frac{u_{0}}{\lambda_{1}}, \quad
    |y_{2}|<|d_{2}|\frac{u_{0}}{\lambda_{2}}$}
\\\\
The boundary of the controllability region $Q$ has two corner points
(see Figure~\ref{fig:Q_Vcontrollabi}):
\begin{equation}\label{eq:boudaries_domai}
   \begin{array}{c}
   y_{1}=-d_{1}\frac{u_{0}}{\lambda_{1}},\quad  y_{2}=-d_{2}\frac{u_{0}}{\lambda_{2}};\\\\
  y_{1}=d_{1}\frac{u_{0}}{\lambda_{1}},\quad  y_{2}=d_{2}\frac{u_{0}}{\lambda_{2}}
\end{array}
\end{equation}
These points (\ref{eq:boudaries_domai}) are the equilibrium points
of  system (\ref{eq:primesjor2}) under the constant controls:
\begin{equation}\label{eq:saturatedcontrol}
    u=\pm u_{0}
\end{equation}

We can ``suppress'' the instability of  the state $y_{1}=0$,
$y_{2}=0$ by a linear feedback control,
\begin{equation}\label{eq:feedbackcont2}
    u=k_{1}y_{1}+k_{2}y_{2}
\end{equation}
with $k_{1}=const$ and $k_{2}=const$.
 It is shown in  paper \cite{HuLinQiu01} that using a linear feedback (\ref{eq:feedbackcont2}) with saturation ($\gamma=const$):
\begin{equation}\label{eq:solrealbi3}
 u=
  \left\{%
\begin{array}{lll}
  u_{0}, & \quad if & \gamma(k_{1}y_{1}+k_{2}y_{2})\geq u_{0} \\
  \\
  \gamma(k_{1}y_{1}+k_{2}y_{2}), & \quad if & |\gamma(k_{1}y_{1}+k_{2}y_{2})|\leq
     u_{0} \\
     \\
  -u_{0}, & \quad if & \gamma(k_{1}y_{1}+k_{2}y_{2}) \leq -u_{0} \\
\end{array}%
\right.
\end{equation}
the basin of attraction $B$  can be made arbitrary close to the
controllability domain $Q$.

The straight line crossing  two points (\ref{eq:boudaries_domai}) is
the following:
\[
    k_{1}y_{1}+k_{2}y_{2}=0
\]
with
\begin{equation}\label{eq:coeffk1k2}
    k_{1}=-\frac{d_{2}}{\lambda_{2}},\quad  k_{2}=\frac{d_{1}}{\lambda_{1}}
\end{equation}
If
\begin{equation}\label{eq:signlambd}
   sign\gamma=sign\left[d_{1}d_{2}\left(\lambda_{1}-\lambda_{2}\right)\right]
\end{equation}
and $|\gamma|\rightarrow\infty$, then the basin of attraction $B$ of
 system (\ref{eq:primesjor2}) under the nonlinear control
(\ref{eq:solrealbi3}) with  coefficients (\ref{eq:coeffk1k2})
tends to the controllability region $Q$. Consequently, using the
coefficients (\ref{eq:coeffk1k2}),  the basin $B$  can be made
arbitrary close to  the domain $Q$. If  $|\gamma|\rightarrow\infty$,  control (\ref{eq:solrealbi3}) tends to the bang-bang control.

The solutions $y_{1}(t)$ and $y_{2}(t)$ of  system (\ref{eq:primesjor2}), (\ref{eq:solrealbi3}) tend to $0$  as $t \rightarrow
     \infty$ for the initial values $y_{1}(0)$, $y_{2}(0)$, belonging to the basin of attraction
     of  system (\ref{eq:primesjor2}),~(\ref{eq:solrealbi3}).
     But if $y_{1}(t) \rightarrow 0$ and $y_{2}(t) \rightarrow 0$, then, according to the expression
(\ref{eq:solrealbi3}), $u(t)\rightarrow 0$ as $t \rightarrow
\infty$. Therefore,  solutions $y_{3}(t)$, $y_{4}(t)$
      of the third and fourth equations of  system (\ref{eq:equajordan1}) with any initial
      conditions $y_{3}(0)$, $y_{4}(0)$
          converge to zero as $t \rightarrow \infty$, because $Re\lambda_{3}<0$, $Re\lambda_{4}<0$.
         Thus, under  control (\ref{eq:solrealbi3}) with  coefficients (\ref{eq:coeffk1k2}), the basin of attraction
of  system
         (\ref{eq:equajordan1}), (\ref{eq:solrealbi3})
         is described by the same relations, which describe the basin of attraction of  system
         (\ref{eq:primesjor2}), (\ref{eq:solrealbi3}).

    The variables $y_{1}$ and $y_{2}$ depend on the original variables from the vector $x$,
    according to  the  transformation $y=S^{-1}x$.
    Due to this,  formula (\ref{eq:solrealbi3}) defines a nonlinear feedback control, which depends on the vector
     $x$ of the original variables. If the matrix $S$ is calculated, then all coefficients of the designed control can be found. Only the constant multiplier $\gamma$ is an arbitrary one; but it has to satisfy  relation (\ref{eq:signlambd}) and to be sufficiently large in modulus.

Thus, linearizing the nonlinear system (\ref{eq:motionequation3}), (\ref{eq:motionequation4}), (\ref{eq:solrealbi3}) near the equilibrium state we get the system, which is asymptotically stable. Then according to Lyapounov's theorem (see \cite{KHALIL02}), the
equilibrium state  (\ref{eq:equilibrium1}) of the nonlinear system
(\ref{eq:motionequation3}), (\ref{eq:motionequation4}) is
asymptotically stable under  control (\ref{eq:solrealbi3}) with
some basin of attraction. In the next Subsection, numerically we find the upper bounds  of the initial values of some variables, which can be handled under designed control.
\subsubsection{Numerical results} Let
\begin{equation}\label{eq:data2}
  \begin{array}{c}
    m_{1}=1.0\thinspace kg,\quad m_{2}=0.2\thinspace kg,\quad g=9.81\thinspace m/s^2,
    \\\\
    r=0.05\thinspace m,\thinspace \thinspace R=0.8\thinspace m,\thinspace \thinspace
     l=0.2\thinspace m,\thinspace \thinspace a=0.15\thinspace m,\thinspace \thinspace
    \rho_{1}=0.2646\thinspace
    m,\thinspace \thinspace \rho_{2}=0.1414\thinspace m\\
  \end{array}
\end{equation}
In open-loop the poles of the linear system (\ref{eq:stateform1}) with  parameters (\ref{eq:data2}) are:
\begin{equation}\label{eq:eigenvalue}
      \lambda_{1}=4.89589,\quad  \lambda_{2}=0.46516,\quad \lambda_{3}=-4.89706,\quad \lambda_{4}=-0.46523\thinspace \\
\end{equation}
Using  formulas (\ref{eq:boudaries_rect}), the controllability
domain $Q$ for  system (\ref{eq:primesjor2}) is designed. It is
bounded in Figure~\ref{fig:Q_Vcontrollabi} by dashed line. Its
boundary contains the corner points (\ref{eq:boudaries_domai}).
Using the linear model (\ref{eq:linear3}), (\ref{eq:linear4}), we
can define the following equilibrium points under  controls
(\ref{eq:saturatedcontrol}) with the original variables:
\begin{equation}\label{eq:controlpointequi}
\begin{array}{c}
    \theta=\mp \frac{c_{u}u_{0}}{g[m_{1}a+m_{2}(l-R)]},\quad\quad \dot{\theta}=0,\\\\ \varphi=-\frac{R}{r}\theta\quad(s=-R\theta),\quad\quad  \dot{\varphi}=0\quad(\dot{s}=0).
    \end{array}
\end{equation}
Points (\ref{eq:controlpointequi}) are located on the boundary
of the controllability region.

Using the nonlinear model (\ref{eq:motionequation3}),
(\ref{eq:motionequation4}), instead of (\ref{eq:controlpointequi})
we get the following expressions:
\begin{equation}\label{eq:controlpointequiarcsin}
\begin{array}{c}
    \theta=\mp arcsin\frac{c_{u}u_{0}}{g[m_{1}a+m_{2}(l-R)]},\quad \quad\dot{\theta}=0,\\\\ \varphi=-\frac{R}{r}\theta\quad(s=-R\theta),\quad \quad\dot{\varphi}=0\quad(\dot{s}=0).
    \end{array}
\end{equation}
In  equilibriums (\ref{eq:controlpointequiarcsin}), the ball is
located on the highest point of the circular beam and consequently
in this point of contact between the ball and the beam the tangent
to the beam is horizontal. Remind, if $u=0$, we have the equilibrium
state (\ref{eq:equilibrium1}).

 The basin of
attraction $B$  for  system (\ref{eq:primesjor2}) under the
control (\ref{eq:solrealbi3}) is shown in the same
Figure~\ref{fig:Q_Vcontrollabi}. Its boundary is drawn in
Figure~\ref{fig:Q_Vcontrollabi} by solid line. This boundary is the
periodical motion (cycle) of   system (\ref{eq:primesjor2}),
(\ref{eq:solrealbi3}).
 This cycle is computed,
using the backward motion of  system (\ref{eq:primesjor2}),
(\ref{eq:solrealbi3}) from a state close to  the origin
$y_{1}=y_{2}=0$. The basin $B$  depends on the coefficient $\gamma$.
We show  in Figure~\ref{fig:Q_Vcontrollabi} the basin of attraction
$B$,  with $\gamma=4000$. If the coefficient $\gamma$ is smaller, then the basin of attraction is smaller too.


In simulation,  the control law (\ref{eq:solrealbi3}) is applied to
the nonlinear model
(\ref{eq:motionequation3}),~(\ref{eq:motionequation4}).
Figure~\ref{fig:phipost2} shows the graphs of the angular variables
$\theta$ and $\varphi$. These graphs are designed for the initial
angle $\varphi(0)=70.39°$ and
$\theta(0)=\dot{\theta}(0)=\dot{\varphi}(0)=0$. This value
$\varphi(0)=70.39°$ is close to the upper bound of the initial
angles $\varphi(0)$, which are possible to  stabilize  the
equilibrium state (\ref{eq:equilibrium1}). The corresponding initial
distance $s(0)$ is equal to $0.061~m$. No oscillations appear during
the transient process, because  matrix $A$ does not have complex
poles.
The voltage, supplied to the motor, is shown in Figure
 \ref{fig:voltage2post2}. The limit value $u_{0}=-19~V$ is
 reached at initial time. No oscillations also appear in
 the graph of  voltage $u(t)$.

The following formula holds for the reaction force $F$, applied to
the ball orthogonally to the beam in their contact point:
\begin{equation}\label{eq:forcecircle}
    F=m_{2}\left[g~cos\left(\theta+\frac{r\varphi}{R}\right)-(R+r)\left(\dot{\theta}+\frac{r\dot{\varphi}}{R}\right)^{2}+(R-l)\dot{\theta}^{2}cos\frac{r\varphi}{R}-(R-l)\ddot{\theta}sin\frac{r\varphi}{R}\right]
\end{equation}
In the numerical experiment, presented in Figures~\ref{fig:phipost2} and \ref{fig:voltage2post2},  force $F$ is always positive. This force is shown in
Figure~\ref{fig:force2}.

Calculating  the values $\theta$ and $s$ with first and third formulas in
(\ref{eq:controlpointequiarcsin}), we obtain
\begin{equation}\label{eq:vanum1}
    \theta=0.469,\quad s=-0.375~m
\end{equation}
Consider the initial velocities $\dot{\theta}(0)=0$, $\dot{s}(0)=0$
and let $s(0)=-R\theta(0)$ be (see third equality in
(\ref{eq:controlpointequiarcsin})). Simulating the nonlinear system
(\ref{eq:motionequation3}), (\ref{eq:motionequation4}) under  control (\ref{eq:solrealbi3}) ($\gamma=4000$), we get values
which are close to the boundary of the attraction basin
\begin{equation}\label{eq:vanum2}
    \theta(0)=0.397,\quad s(0)=-0.318~m
\end{equation}
These values (\ref{eq:vanum2}) equal to values (\ref{eq:vanum1})
divided by $1.18$. Under the nonlinear control law (\ref{eq:solrealbi3}) with
$\gamma=8000$ we come to the values\\
\[
    \theta(0)=0.430,\quad s(0)=-0.344~m
\]\\
These initial conditions equal to values (\ref{eq:vanum1}) divided
by $1.09$ and they are closer to (\ref{eq:vanum1}) than values
(\ref{eq:vanum2}). Our numerical experiments show that possible for
stabilization initial values $\theta(0)$, $s(0)$ tend to  values
(\ref{eq:vanum1}) as $\gamma\rightarrow\infty$. Thus, formulas
(\ref{eq:controlpointequiarcsin}) can be used to evaluate the basin
of attraction for the original nonlinear system
(\ref{eq:motionequation3}), (\ref{eq:motionequation4}).

We think there is no admissible control $|u(x)| \leq u_{0}$ to bring  system (\ref{eq:motionequation3}), (\ref{eq:motionequation4}) to equilibrium (\ref{eq:equilibrium1}) from the initial states

\[
\begin{array}{c}
     \dot{\theta}(0)=\dot{\varphi}(0)=0,\quad |\theta(0)|\geq arcsin\frac{c_{u}u_{0}}{g[m_{1}a+m_{2}(l-R)]}, \\\\ \varphi(0)=-\frac{R}{r}\theta(0)\quad(s(0)=-R\theta(0)).
     \end{array}
\]\\
This hypothesis is similar to the corresponding hypothesis for the straight beam-and-ball system (see subsection~\ref{sectionnum}).

\section{Conclusion}\label{sec:conclusion}
In this article, we consider the well known straight beam-and-ball
system and an original circular beam-and-ball system. The problem of
stabilization of unstable equilibriums of these systems is studied.
The model linearized near the unstable equilibrium of the straight
beam-and-ball system has one unstable mode. The difficulty is
greater to stabilize the circular beam-and-ball system, because its
linear model has two unstable modes. For each system we use the
Jordan form of the linear model to extract the unstable part and to
stabilize the equilibrium. Considering the restriction on the
voltage of the motor, the objective is to get a large basin of
attraction. The designed feedback control contains the unstable Jordan variables only. All parameters of this control are defined up to a constant multiplier. Simulation results
for the complete nonlinear systems are shown. These results are
close for linear and nonlinear systems. All the numerical results,
obtained in this paper
 for nonlinear systems, are
realistic and illustrate the efficiency of the designed control
laws. Using the described above approach for both cases it is easily
to take into account the friction forces also. Testbed devices can
be now imagined
 to test the designed control laws experimentally. The
original circular beam-and-ball system will be interesting for
demonstrations, devoted to education and to investigate new
nonlinear control laws.
\section{List of Captions}\label{sec:caption}
\begin{figure}[H] 
    \centerline{\includegraphics[width=14cm]{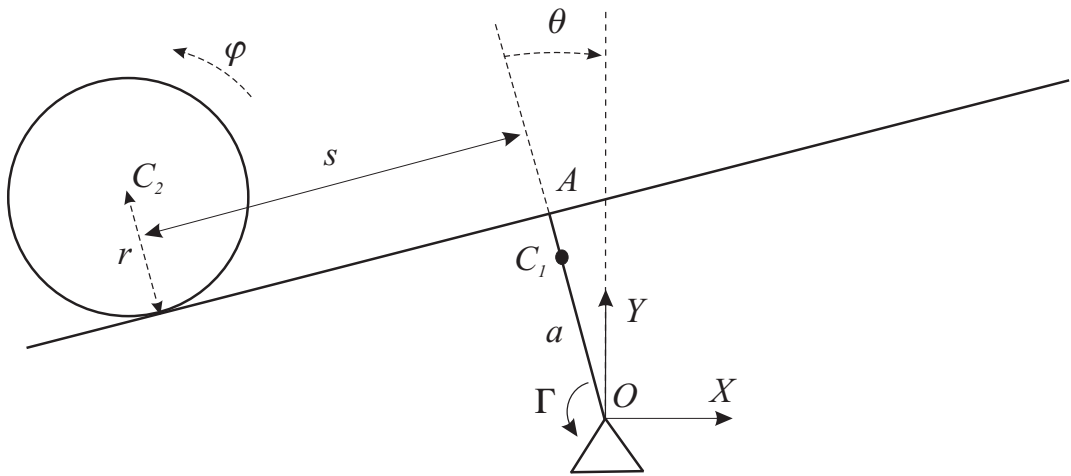}}
    \centerline{} \caption[]{Diagram of the straight beam-and-ball system.}
    \label{fig:diagram1}
  \end{figure}

\begin{figure}[H] 
\begin{center}
         \psfrag{t(s)}{\qquad \qquad \qquad \quad  \tiny{Time  [s]}}
          \psfrag{theta}{\footnotesize${\theta(t)}$}
                 \psfrag{phi}{\footnotesize{${\varphi(t)}$}}
       \centerline{\epsfig{file = 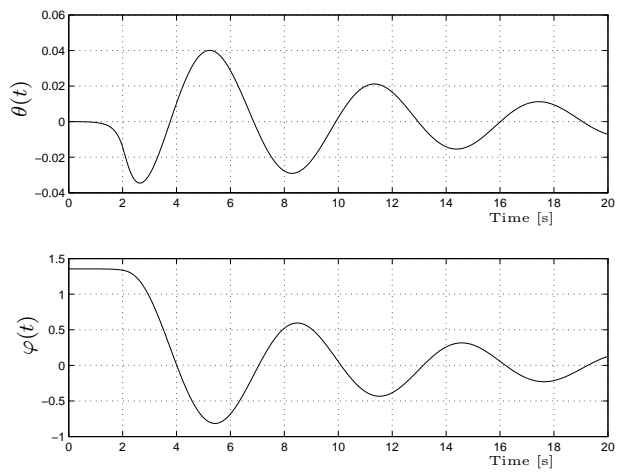,scale = 0.45}}
       \caption[]{Stabilization of the Straight beam-and-ball system, $\theta\rightarrow0$ and $\varphi\rightarrow0$ (in radians).}
       \label{fig:phipost1}
\end{center}
\end{figure}

\begin{figure}[H] 
\begin{center}
               \psfrag{t(s)}{\qquad \qquad \qquad \quad \tiny{Time  [s]}}
         \psfrag{Voltage}{\qquad \qquad \quad \footnotesize{$Volts$}}
            \psfrag{U(t)}{$U(t)$}
          \centerline{\epsfig{file = 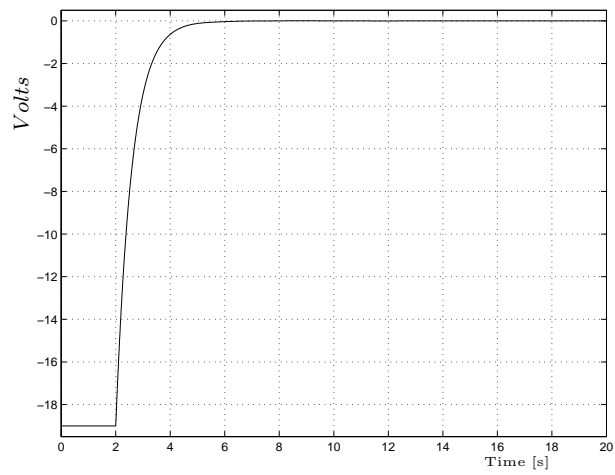,scale = 0.455}}
            \caption[]{Voltage $u(t)$, supplied to the motor for the stabilization  of the Straight beam-and-ball system.}
       \label{fig:voltage1post1}
\end{center}
\end{figure}

\begin{figure}[H] 
\begin{center}
               \psfrag{t(s)}{\qquad \qquad \qquad \quad \tiny{Time  [s]}}
           \psfrag{N}{\qquad \qquad \quad \footnotesize{$N$}}
          \centerline{\epsfig{file = 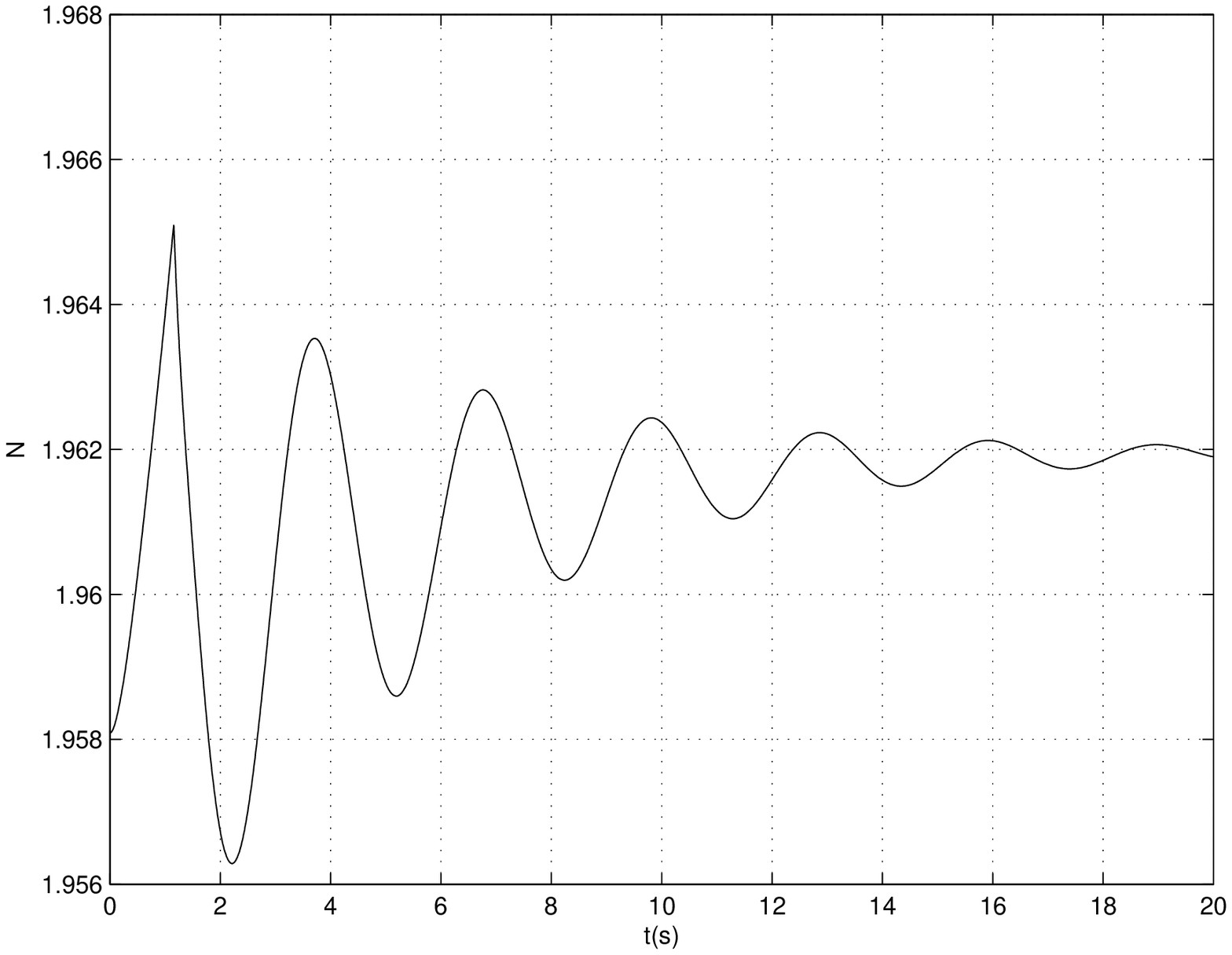,scale = 0.455}}
            \caption[]{The reaction force $F(t)$, applied to the ball during the stabilization process.}
       \label{fig:force1}
\end{center}
\end{figure}

\begin{figure}[H] 
    \centerline{\includegraphics[width=14cm]{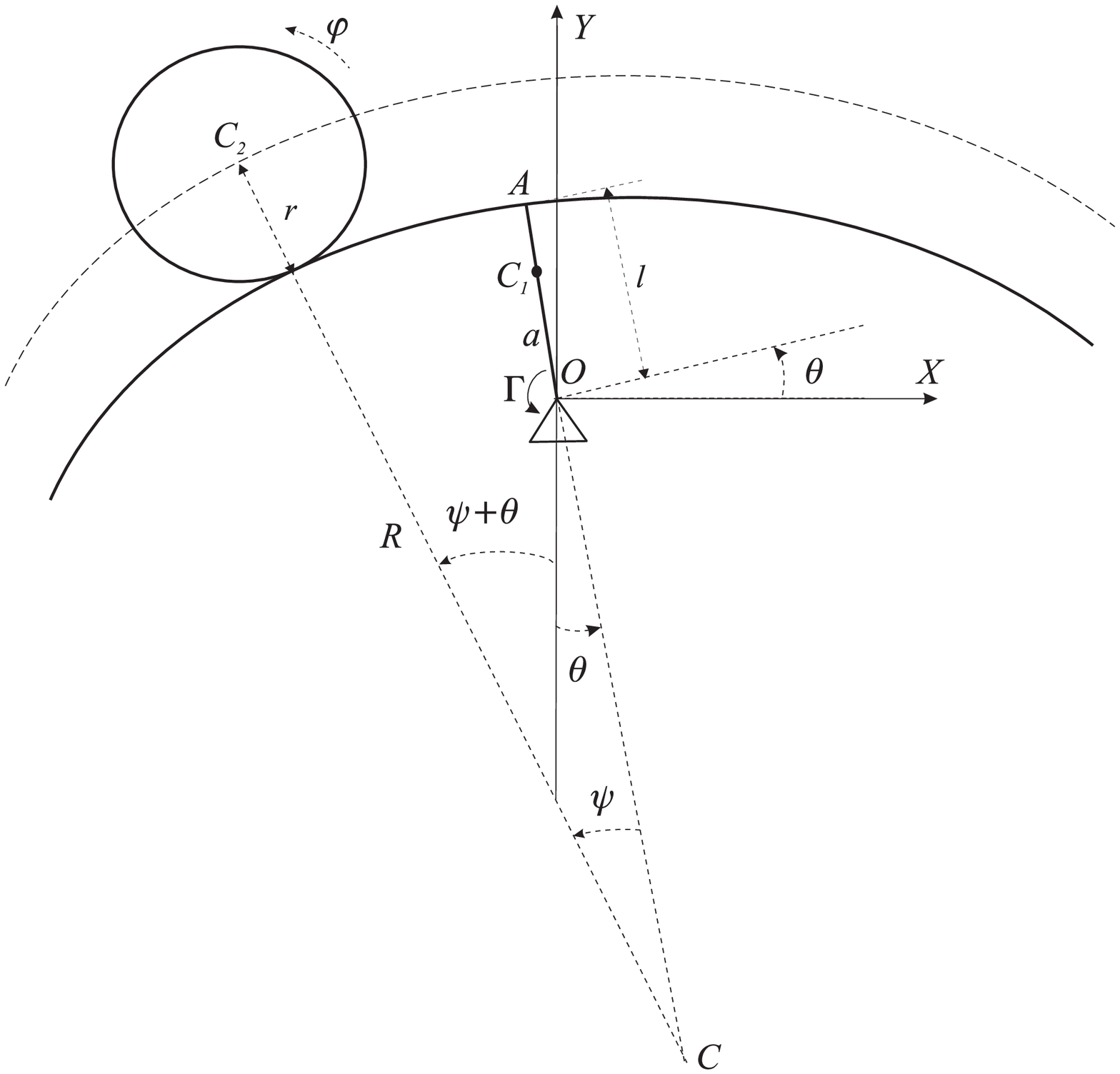}}
    \centerline{} \caption[]{Diagram of the circular beam-and-ball system.}
    \label{fig:diagram2}
  \end{figure}

\begin{figure}[H] 
\begin{center}
      \psfrag{y1}{$
\begin{array}{*{1}c}
   {} \\
   {y_1}\\
\end{array}$}
      \psfrag{y2}{$y_{2}$}
       \centerline{\epsfig{file = 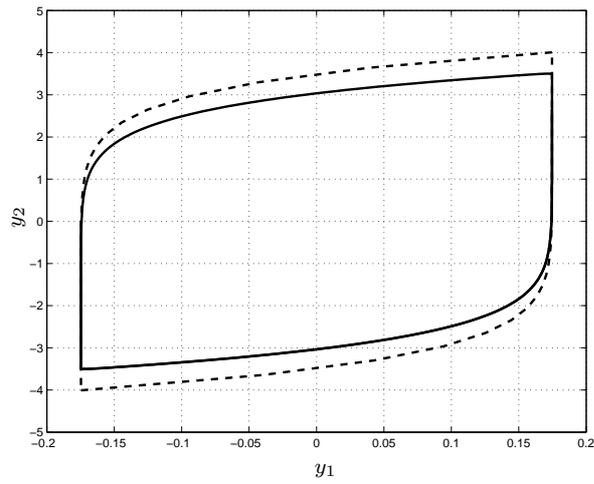,scale = 0.45}}
       \caption[]{Controllability domain $Q$ (dashed line) for  system (\ref{eq:primesjor2}) and basin of attraction $B$  (solid line) with $\gamma=4000$}.
       \label{fig:Q_Vcontrollabi}
\end{center}
\end{figure}

\begin{figure}[H] 
\begin{center}
         \psfrag{t(s)}{\qquad \qquad \qquad \quad  \tiny{Time  [s]}}
          \psfrag{theta}{\footnotesize${\theta(t)}$}
                 \psfrag{phi}{\footnotesize{${\varphi(t)}$}}
       \centerline{\epsfig{file = 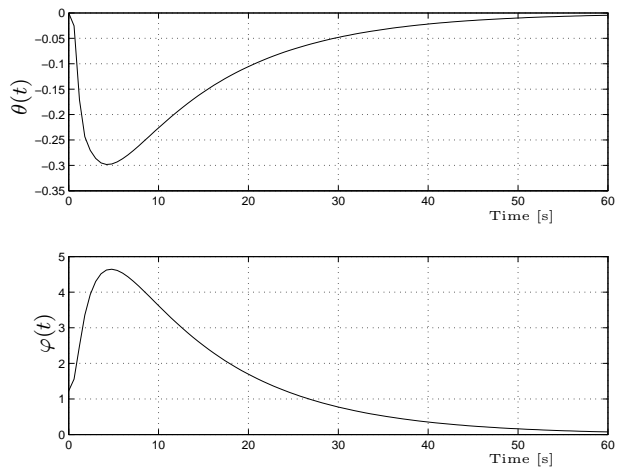,scale = 0.45}}
       \caption[]{Stabilization of the Circular beam-and-ball system, $\theta\rightarrow0$ and $\varphi\rightarrow0$ (in radians).}
       \label{fig:phipost2}
\end{center}
\end{figure}

\begin{figure}[H] 
\begin{center}
               \psfrag{t(s)}{\qquad \qquad \qquad \quad \tiny{Time  [s]}}
          \psfrag{Voltage}{\qquad \qquad \quad \footnotesize{$Volts$}}
            \psfrag{u(t)}{$u(t)$}
          \centerline{\epsfig{file = 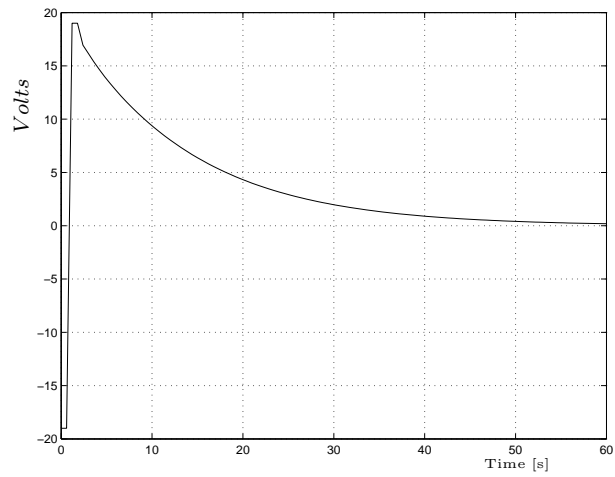,scale = 0.455}}
            \caption[]{Voltage $u(t)$, supplied to the motor for the stabilization  of the Circular beam-and-ball system.}
       \label{fig:voltage2post2}
\end{center}
\end{figure}

\begin{figure}[H] 
\begin{center}
               \psfrag{t(s)}{\qquad \qquad \qquad \quad \tiny{Time  [s]}}
            \psfrag{N}{\qquad \qquad \quad \footnotesize{$N$}}
          \centerline{\epsfig{file = 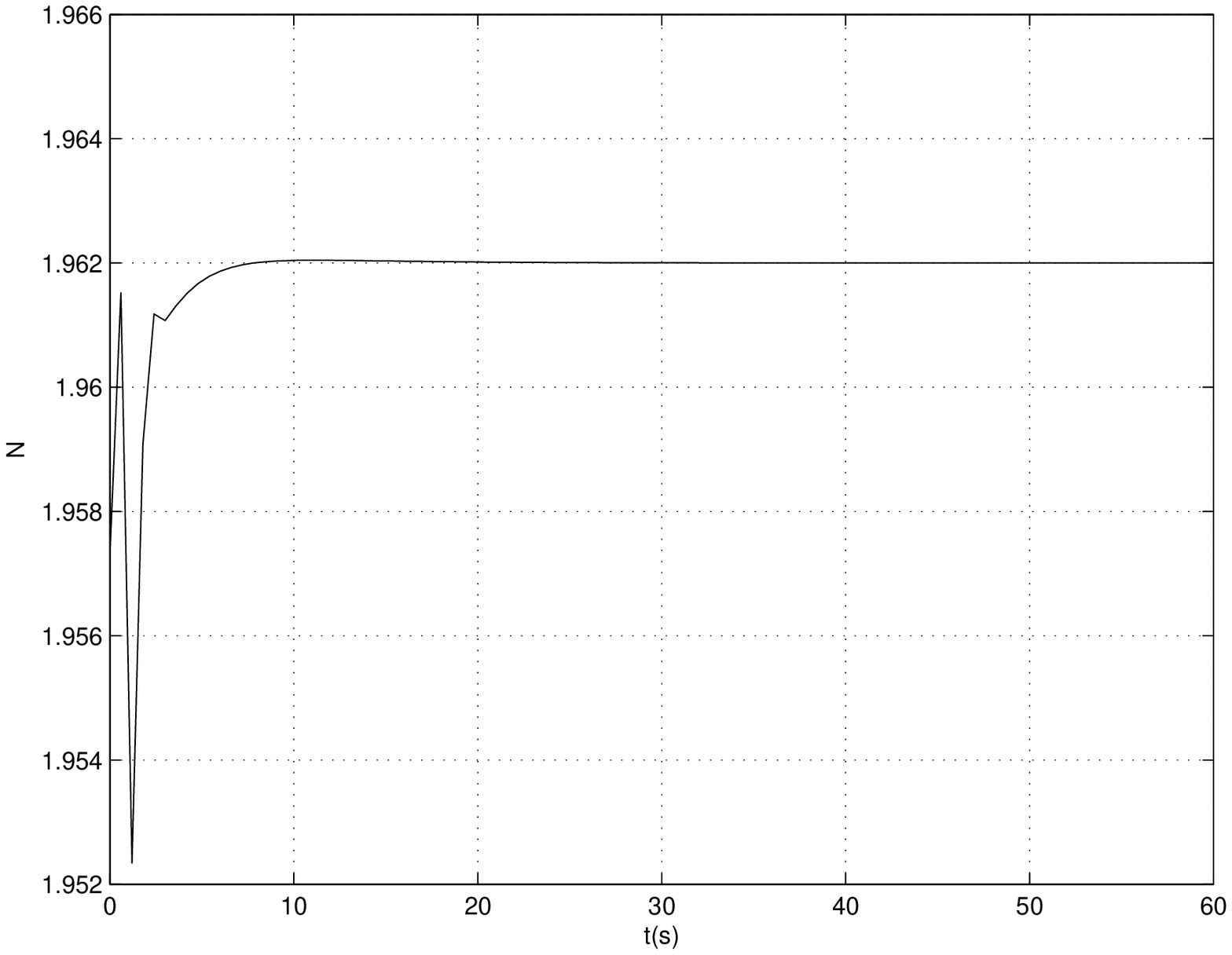,scale = 0.455}}
            \caption[]{The reaction force $F(t)$, applied to the ball during the stabilization process.}
       \label{fig:force2}
\end{center}
\end{figure}

\begin{enumerate}
  \item Figure \ref{fig:diagram1}: Diagram of the straight beam-and-ball system
  \item Figure \ref{fig:phipost1}: Stabilization of the Straight beam-and-ball system, $\theta\rightarrow0$ and $\varphi\rightarrow0$ (in radians).
  \item Figure \ref{fig:voltage1post1}: Voltage $u(t)$, supplied to the motor for the stabilization  of the Straight beam-and-ball system.
  \item Figure \ref{fig:force1}: The reaction force $F(t)$, applied to the ball during the stabilization process.
  \item Figure \ref{fig:diagram2}: Diagram of the circular beam-and-ball system.
  \item Figure \ref{fig:Q_Vcontrollabi}: Controllability domain $Q$ (dashed line) for  system (\ref{eq:primesjor2}) and basin of attraction $B$  (solid line) with
  $\gamma=4000$.
  \item Figure \ref{fig:phipost2}: Stabilization of the Circular beam-and-ball system,
$\theta\rightarrow0$ and $\varphi\rightarrow0$ (in radians).
  \item Figure \ref{fig:voltage2post2}: Voltage $u(t)$, supplied to the motor for the stabilization  of the circular beam-and-ball system.
  \item Figure \ref{fig:force2}: The reaction force $F(t)$, applied to the ball during the stabilization process.
\end{enumerate}

\bibliographystyle{IEEEtran}
\bibliography{biblio}
\vspace{12pt}

\end{document}